\documentclass[letterpaper]{article} 
\usepackage{aaai24}  
\usepackage{times}  
\usepackage{helvet}  
\usepackage{courier}  
\usepackage[hyphens]{url}  
\usepackage{graphicx} 
\usepackage{natbib}  
\usepackage{caption} 
\usepackage{algorithm}
\usepackage{algorithmic}

\usepackage{newfloat}
\usepackage{listings}
\DeclareCaptionStyle{ruled}{labelfont=normalfont,labelsep=colon,strut=off} 
\floatstyle{ruled}
\newfloat{listing}{tb}{lst}{}
\floatname{listing}{Listing}
\usepackage{amsmath}
\usepackage{amssymb}
\usepackage{mathtools}
\usepackage{amsthm}
\usepackage{booktabs}
\usepackage{hyperref}
\title{Scaling Transformer to 1M tokens and beyond with RMT}
\author{
    Aydar Bulatov\textsuperscript{\rm 1}
    ,
    Yuri Kuratov\textsuperscript{\rm 2,1},
    Yermek Kapushev\textsuperscript{\rm 2},
    Mikhail Burtsev\textsuperscript{\rm 3}
}
\affiliations{
    \textsuperscript{\rm 1}Neural Networks and Deep Learning Lab, MIPT, Dolgoprudny, Russia \\
    \textsuperscript{\rm 2}AIRI, Moscow, Russia \\
    \textsuperscript{\rm 3}London Institute for Mathematical Sciences, London, UK\\

    \{bulatov.as,yurii.kuratov\}@phystech.edu, mb@lims.ac.uk
}

\begin{document}

\maketitle

\begin{abstract}
A major limitation for the broader scope of problems solvable by transformers is the quadratic scaling of computational complexity with input size. In this study, we investigate the recurrent memory augmentation of pre-trained transformer models to extend input context length while linearly scaling compute. Our approach demonstrates the capability to store information in memory for sequences of up to an unprecedented two million tokens while maintaining high retrieval accuracy. Experiments with language modeling tasks show perplexity improvement as the number of processed input segments increases. These results underscore the effectiveness of our method, which has significant potential to enhance long-term dependency handling in natural language understanding and generation tasks, as well as enable large-scale context processing for memory-intensive applications.
\end{abstract}

\section{Introduction}

Transformer-based models show their effectiveness across multiple domains and tasks. The self-attention allows to combine information from all sequence elements into context-aware representations. However, global and local information has to be stored mostly in the same element-wise representations. Moreover, the length of an input sequence is limited by quadratic computational complexity of self-attention.

In this work, we propose and study a memory-augmented segment-level recurrent Transformer (Recurrent Memory Transformer or RMT). Memory allows to store and process local and global information as well as to pass information between segments of the long sequence with the help of recurrence.
We implement a memory mechanism with no changes to Transformer model by adding special memory tokens to the input or output sequence. Then Transformer is trained to control both memory operations and sequence representations processing.

In this study we show that by using simple token-based memory mechanism introduced in~\citep{rmt_2022} can be combined with pretrained transformer models like BERT~\citep{devlin2019bert} and GPT-2~\citep{radford2019gpt2} with full attention and full precision operations.

\paragraph{Contributions}
1.  We expand application of RMT to encoder-only and decoder-only pre-trained language models. Proposed segment wise curriculum learning allows to fine-tune majority of pre-trained transformer based models for processing potentially unlimited sequences. 

2. To benchmark generalization capabilities of RMT we propose a set of novel memory acquisition and retention tasks scalable to extremely long sequences of million tokens.

3. We demonstrate the unparalleled ability of RMT to generalize memory operations, successfully detecting and storing information about facts for up to two million tokens. To the best of our knowledge, this establishes a record for the longest sequence task processed by any existing deep neural network. Furthermore, we identify no technical limitations that prevent further scaling.

4. We compare computational complexity of RMT vs. other transformer models and demonstrate the significant advantage of RMT due to its linear scaling of inference operations and constant memory.

The code is available on GitHub\footnote{\url{https://github.com/booydar/t5-experiments/tree/aaai24}}.
The paper version with supplementary materials is available on arXiv\footnote{\url{https://arxiv.org/abs/2304.11062}}.

\section{Related Work}

Our work revolves around the concept of memory in neural architectures. Memory has been a recurrent theme in neural network research, dating back to early works~\citep{mcculloch1943logical,stephen1956kleene} and significantly advancing in the 1990s with the introduction of the \textit{Backpropagation Through Time} learning algorithm~\citep{werbos1990backpropagation} and \textit{Long-Short Term Memory} (LSTM) neural architecture~\citep{10.1162/neco.1997.9.8.1735}. Contemporary memory-augmented neural networks (MANNs) typically utilize some form of recurrent external memory separate from the model's parameters. \textit{Neural Turing Machines} (NTMs)~\citep{graves2014neural} and \textit{Memory Networks}~\citep{weston2014memory} are equipped with storage for vector representations accessible through an attention mechanism. Memory Networks~\citep{weston2014memory,sukhbaatar2015endtoend} were designed to enable reasoning through sequential attention over memory content.

NTMs, followed by \textit{Differentiable Neural Computer} (DNC)~\citep{graves2016hybrid} and \textit{Sparse DNC}~\citep{rae2016scaling}, are implemented as recurrent neural networks capable of writing to memory storage over time. All these models are differentiable and trainable via backpropagation through time (BPTT). Parallel research lines extend recurrent neural networks, such as LSTM, with data structures like stacks, lists, or queues~\citep{arm2015inferring,grefenstette2015learning}. MANN architectures with more advanced addressing mechanisms, such as address-content separation and multi-step addressing, have been proposed in~\citep{gulcehre2016dynamic,gulcehre2017memory,meng2018context}. The Global Context Layer model~\citep{meng2018context} employs address-content separation to address the challenge of training content-based addressing in canonical NTMs.

Memory is often combined with Transformers in a recurrent approach. Long inputs are divided into smaller segments, processed sequentially with memory to access information from past segments. Transformer-XL~\citep{dai2019transformerxl} preserves previous hidden states for reuse in subsequent segments, while Compressive Transformer~\citep{rae2019compressive} adds new compressed memory. Ernie-Doc~\citep{ding-etal-2021-ernie-doc} enhances contextual information flow by employing same-layer recurrence instead of attending to previous layer outputs of preceding segments. Memformer~\citep{wu2020memformer} introduces a dedicated memory module to store previous hidden states in summarized representations. Using a similar approach to Memformer, MART~\citep{lei2020mart} and Block-Recurrent Transformer~\citep{hutchins2022blockrecurrent} adopt memory update rules analogous to LSTM~\citep{10.1162/neco.1997.9.8.1735} and GRU~\citep{cho2014gru}. FeedBack Transformer~\citep{fan2020feedback-transformer} implements full recurrence beyond the segment level and merges low and high layers representations into a memory state.

A drawback of most existing recurrent methods is the need for architectural modifications that complicate their application to various pre-trained models. In contrast, the Recurrent Memory Transformer can be built upon any model that uses a common supported interface.

Some approaches redesign the self-attention mechanism to reduce computational complexity while minimizing input coverage loss. \textit{Star-Transformer}~\citep{guo2019startransformer}, \textit{Longformer}~\citep{beltagy2020longformer}, \textit{GMAT}~\citep{gupta2020gmat}, \textit{Extended Transformer Construction} (ETC)~\citep{ainslie2020encoding}, and \textit{Big Bird}~\citep{zaheer2020big} limit attention distance and employ techniques such as global representations to preserve long-range dependencies. \textit{Memory Transformer}~\citep{burtsev2020memory-transformer} introduces memory by extending the unchanged model input with special memory tokens.

A common constraint of these methods is that memory requirements grow with input size during both training and inference, inevitably limiting input scaling due to hardware constraints. In contrast, recurrent approaches have constant memory complexity during inference. The longest Longformer, Big Bird, and Long T5~\citep{guo-etal-2022-longt5}, LongNet~\citep{ding2023longnet} models reported in their respective papers have a maximum length of less than 33,000 tokens. CoLT5~\citep{ainslie2023colt5} can handle up to 64,000 tokens. LongNet with sparse delated attention can potentially handle up to 1B tokens per batch in distributed setting. However, CoLT5 and LongNet hit the GPU limit and run out of memory with longer inputs. Furthermore, LongNet is unable to scale to millions of tokens on single GPU due to memory constraints~\citep{ding2023longnet}. Memorizing Transformers~\citep{wu2022memorizing} and Unlimiformer~\citep{bertsch2023unlimiformer} further extend memory through k-NN.

Another line of recent related work focuses on models with an alternative to traditional attention mechanism: S4~\citep{gu2021s4}, Hyena~\citep{poli2023hyena}, RWKV~\citep{peng2023rwkv}, RetNet~\citep{sun2023retentive}. They aim to combine the best of convolutions and recurrence -- high parallelism during training and linear scaling with sequence length. As architectures change, these methods require training models from scratch and are unable to reuse plethora of already pre-trained transformer-based models.

\section{Recurrent Memory Transformer}

Starting from the initial Recurrent Memory Transformer~\citep{rmt_2022} (RMT), we adapted it for a plug-and-play approach as a wrapper for a range of popular Transformers. 

\begin{figure}[t]
\includegraphics[width=\linewidth]{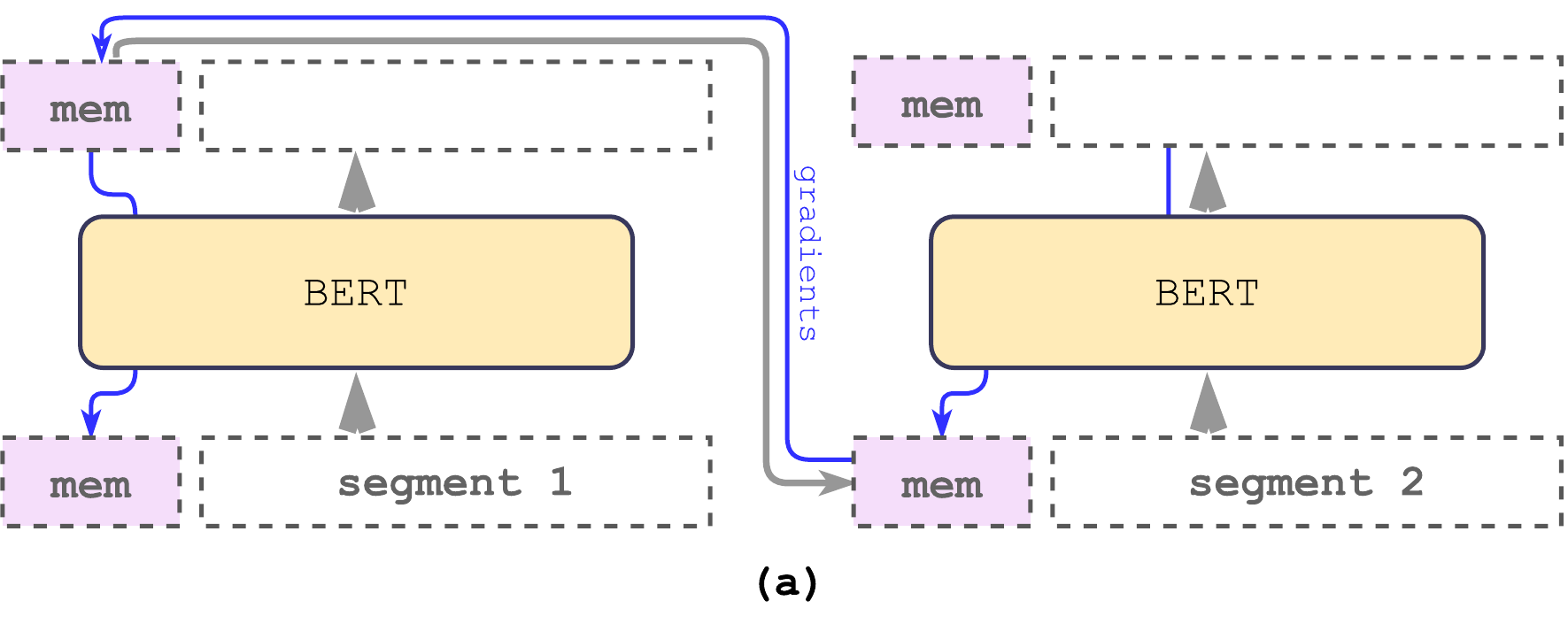}
\includegraphics[width=\linewidth]{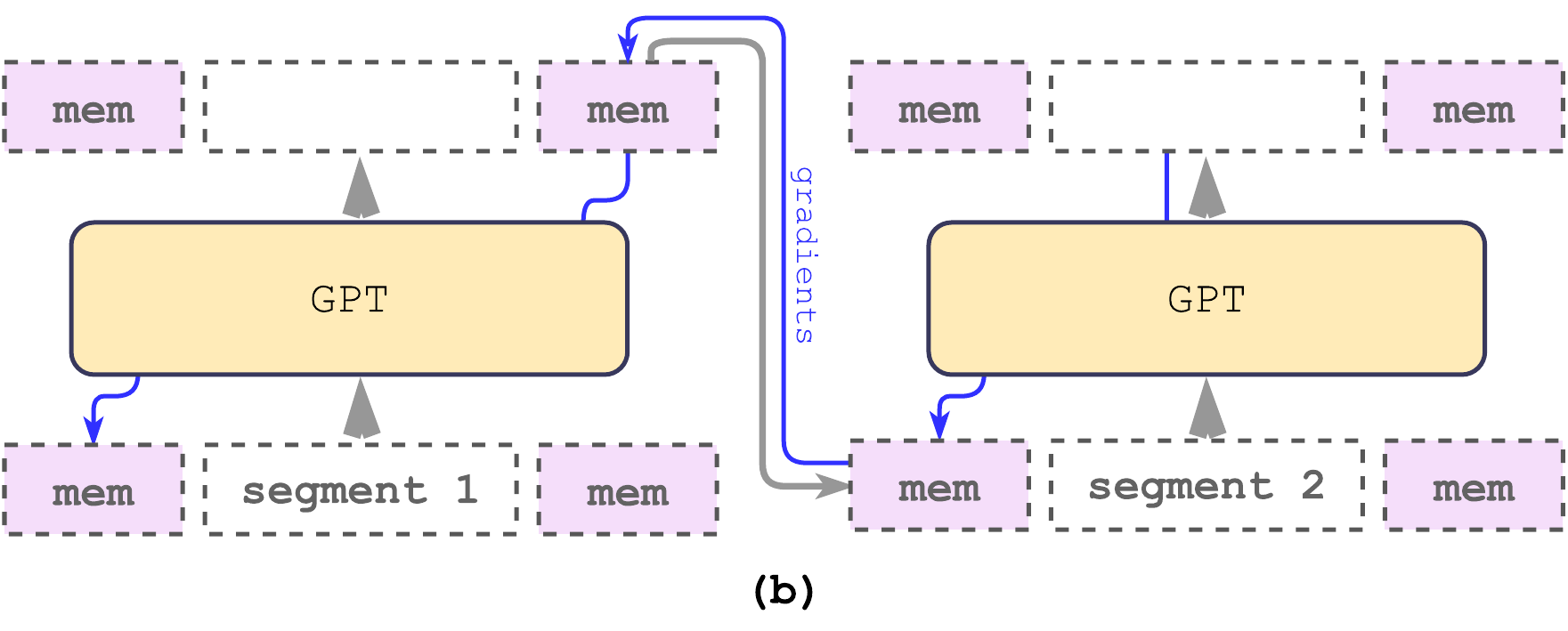}
\caption{\small\textbf{Recurrent memory mechanism.} Memory is passed to Transformer along input sequence embeddings, and memory output is passed to the next segment. \textbf{(a)} For encoder-only models there is a single memory and \textbf{(b)} for decoder models with causal attention mask we add an additional write memory at the end. During training gradients flow from the current segment through memory to the previous segment.}
\label{RMT_simple}
\end{figure}

\begin{figure*}[h]
\centering
\includegraphics[width=0.8\linewidth]{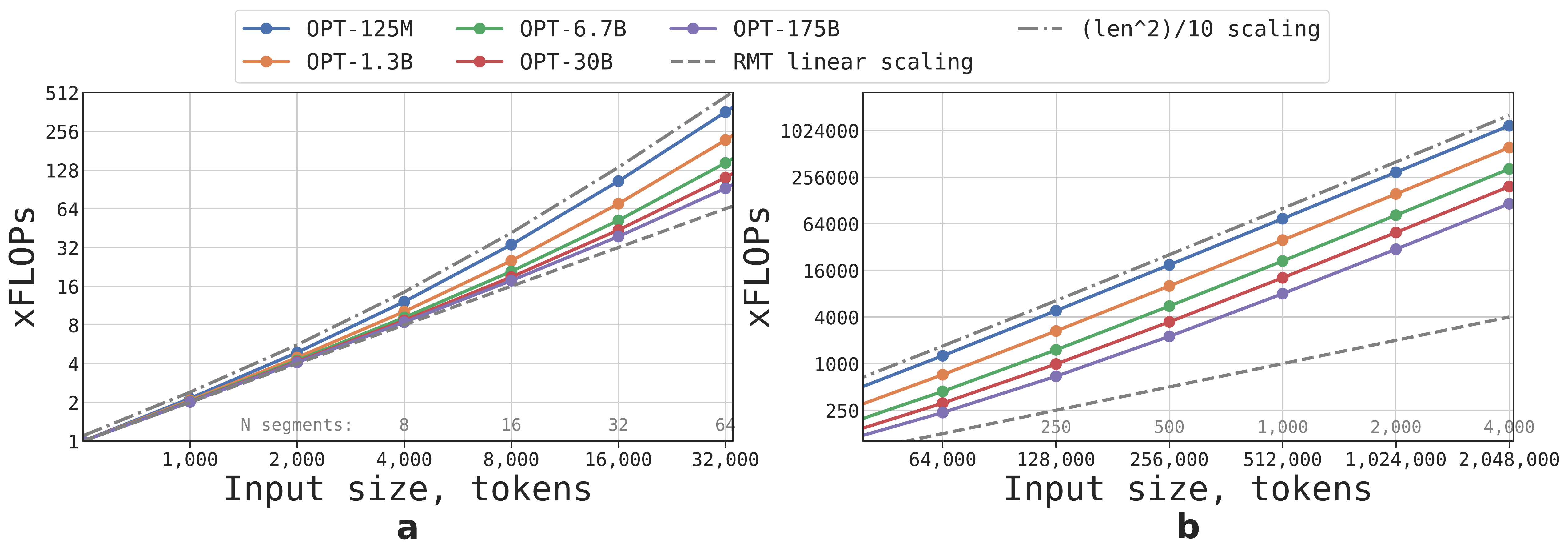}

\caption{\small\textbf{RMT inference scales linearly with respect to the input sequence length}. We estimate the required FLOP increase for the forward pass compared to running models on sequences with 512 tokens.  \textbf{a}: lengths from $512$ to $32{,}000$ tokens, \textbf{b}: lengths from $32{,}000$ to $2{,}048{,}000$ tokens. The RMT segment length is fixed at 512 tokens. While larger models (OPT-30B, OPT-175B) tend to exhibit near-linear scaling on relatively short sequences up to $32{,}000$, they reach quadratic scaling on longer sequences. The angles in the log-log scale indicate a power of a polynomial function. Full attention transformer-based OPT models are much closer to quadratic scaling (dash-dot line). Smaller models (OPT-125M, OPT-1.3B) demonstrate quadratic scaling even on shorter sequences. On sequences with $2{,}048{,}000$ tokens, RMT can run OPT-175B with $\times29$ fewer FLOPs and with $\times295$ fewer FLOPs than OPT-135M.}
\label{fig:complexity_scale}

\end{figure*}

This adaptation augments its backbone with memory, composed of $m$ real-valued trainable vectors (Figure~\ref{RMT_simple}). The lengthy input is divided into segments, and memory vectors are prepended to the first segment embeddings and processed alongside the segment tokens. For encoder-only models like BERT, memory is added only once at the beginning of the segment, unlike~\citep{rmt_2022}, where decoder-only models separate memory into read and write sections. For the time step $\tau$ and segment $H_{\tau}^0$, the recurrent step is performed as follows:

\begin{equation*}
\begin{gathered}
\tilde{H}^{0}_{\tau} = [H_{\tau}^{mem} \circ H_{\tau}^0],
\bar{H}^N_\tau = \text{Transformer}(\tilde{H}^{0}_{\tau}),\\
[ \bar H_{\tau}^{mem} \circ H_{\tau}^N] := \bar{H}^N_\tau,
\end{gathered}
\end{equation*}
here $N$ is a number of Transformer layers.

After the forward pass, $ \bar H_{\tau}^{mem}$ contains updated memory tokens for the segment $\tau$.

Segments of the input sequence are processed sequentially. To enable the recurrent connection, we pass the outputs of the memory tokens from the current segment to the input of the next one:
\begin{equation*}
\begin{gathered}
H_{\tau+1}^{mem} := \bar H_{\tau}^{mem},
\tilde{H}^{0}_{\tau+1} = [H_{\tau+1}^{mem} \circ H_{\tau+1}^0 ].
\end{gathered}
\end{equation*}

Both memory and recurrence in the RMT are based only on global memory tokens. This allows the backbone Transformer to remain unchanged, making the RMT memory augmentation compatible with any Transformer-based model.

We can estimate the required FLOPs for RMT and Transformer models of different sizes and sequence lengths. We took configurations (vocabulary size, number of layers, hidden size, intermediate hidden size, and number of attention heads) for the OPT model family~\citep{Zhang2022OPTOP} and computed the number of FLOPs for the forward pass following~\citep{chinchilla}. We also modified FLOP estimates to account for the effect of RMT recurrence.

Figure~\ref{fig:complexity_scale} shows that RMT scales linearly for any model size if the segment length is fixed. We achieve linear scaling by dividing an input sequence into segments and computing the full attention matrix only within segment boundaries. Larger Transformer models tend to exhibit slower quadratic scaling with respect to sequence length because of compute-heavy FFN layers (which scale quadratically with respect to hidden size). However, on extremely long sequences $>32{,}000$, they fall back to quadratic scaling. RMT requires fewer FLOPs than non-recurrent models for sequences with more than one segment ($>512$ in this study) and can reduce the number of FLOPs by up to $\times295$ times. RMT provides a larger relative reduction in FLOPs for smaller models, but in absolute numbers, a $\times29$ times reduction for OPT-175B models is highly significant. Additional experimental comparison of computational efficiency can be found in the Appendix.

\section{Memorization Tasks}

To test memorization abilities, we constructed synthetic datasets that require memorization of simple facts and basic reasoning. The task input consists of one or several facts and a question that can be answered only by using all of these facts. To increase the task difficulty, we added natural language text unrelated to the questions or answers. This text acts as noise, so the model's task is to separate facts from irrelevant text and use them to answer the questions. The task is formulated as a multi-class classification, with each class representing a separate answer option.

Facts are generated using the bAbI dataset~\citep{WestonBCM15}, while the background text is sourced from questions in the QuALITY~\citep{pang-etal-2022-quality} long QA dataset.

Background text example:  " \textit{... He was a big man,
broad-shouldered and still thin-waisted. 
Eddie found it easy to believe the
stories he had heard about his father ...} "

\begin{figure}
\centering
\includegraphics[width=0.99\linewidth]{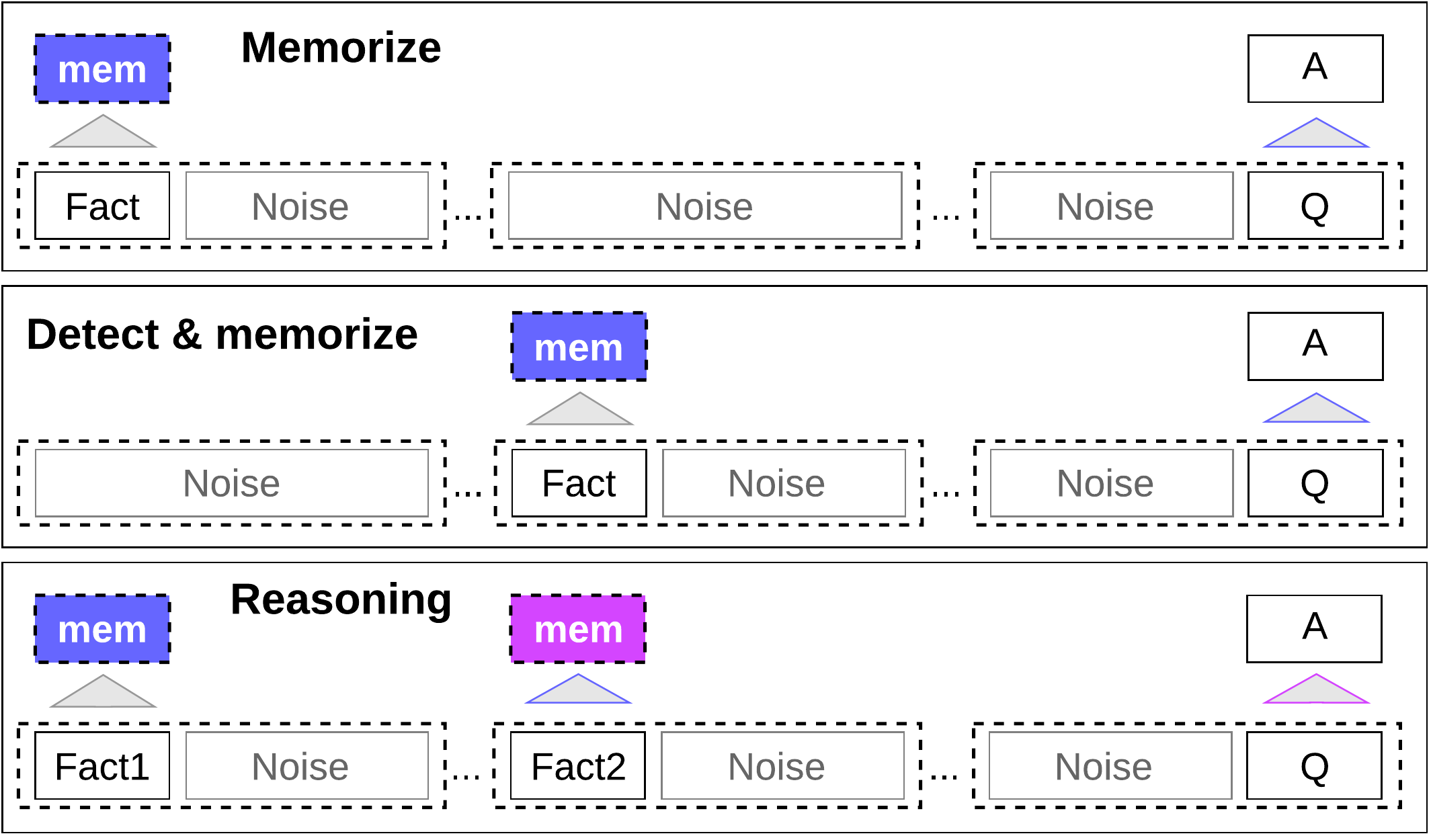}
\caption{\small\textbf{Memory-intensive synthetic tasks}. Synthetic tasks and the required RMT operations to solve them are presented. In the Memorize task, a fact statement is placed at the start of the sequence. In the Detect and Memorize task, a fact is randomly placed within a text sequence, making its detection more challenging. In the Reasoning task, two facts required to provide an answer are randomly placed within the text. For all tasks, the question is at the end of the sequence. 'mem' denotes memory tokens, 'Q' represents the question, and 'A' signifies the answer.}
\label{tasks-approaches}
\end{figure}

The first task tests the ability of RMT to write and store information in memory for an extended time (Figure~\ref{tasks-approaches}, top). In the simplest case, the fact is always located at the beginning of the input, and the question is always at the end. The amount of irrelevant text between the question and answer is gradually increased, so that the entire input does not fit into a single model input. Example: "\textit{Fact: Daniel went back to the hallway. Question: Where is Daniel? Answer: hallway}"

Fact detection increases the task difficulty by moving the fact to a random position in the input (Figure~\ref{tasks-approaches}, middle). This requires the model to first distinguish the fact from irrelevant text, write it to memory, and later use it to answer the question located at the end.

Another important operation with memory is being able to operate with several facts and current context. To evaluate this function, we use a more complicated task called "reasoning",
where two facts are generated and positioned randomly within the input sequence (Figure~\ref{tasks-approaches}, bottom). The question posed at the end of the sequence is formulated in a way that any of the facts must be used to answer the question correctly (i.e., the \textit{Two Argument Relation} bAbI task). Example: "\textit{Fact1: The hallway is east of the bathroom. Fact2: The bedroom is west of the bathroom. Question: What is the bathroom east of? Answer: bedroom}"

\section{Learning Memory Operations}

We use the pretrained models from Hugging Face Transformers~\citep{wolf2020transformers} as backbones for RMT in our experiments. All models are augmented with memory and trained using the AdamW optimizer~\citep{loshchilov2018adamw} with linear learning rate scheduling and warmup.
The technical details of training and the full set of hyperparameters are available in the Appendix and training scripts in the GitHub repository.

\begin{figure}[!t]
\centering
\includegraphics[width=0.8\linewidth]{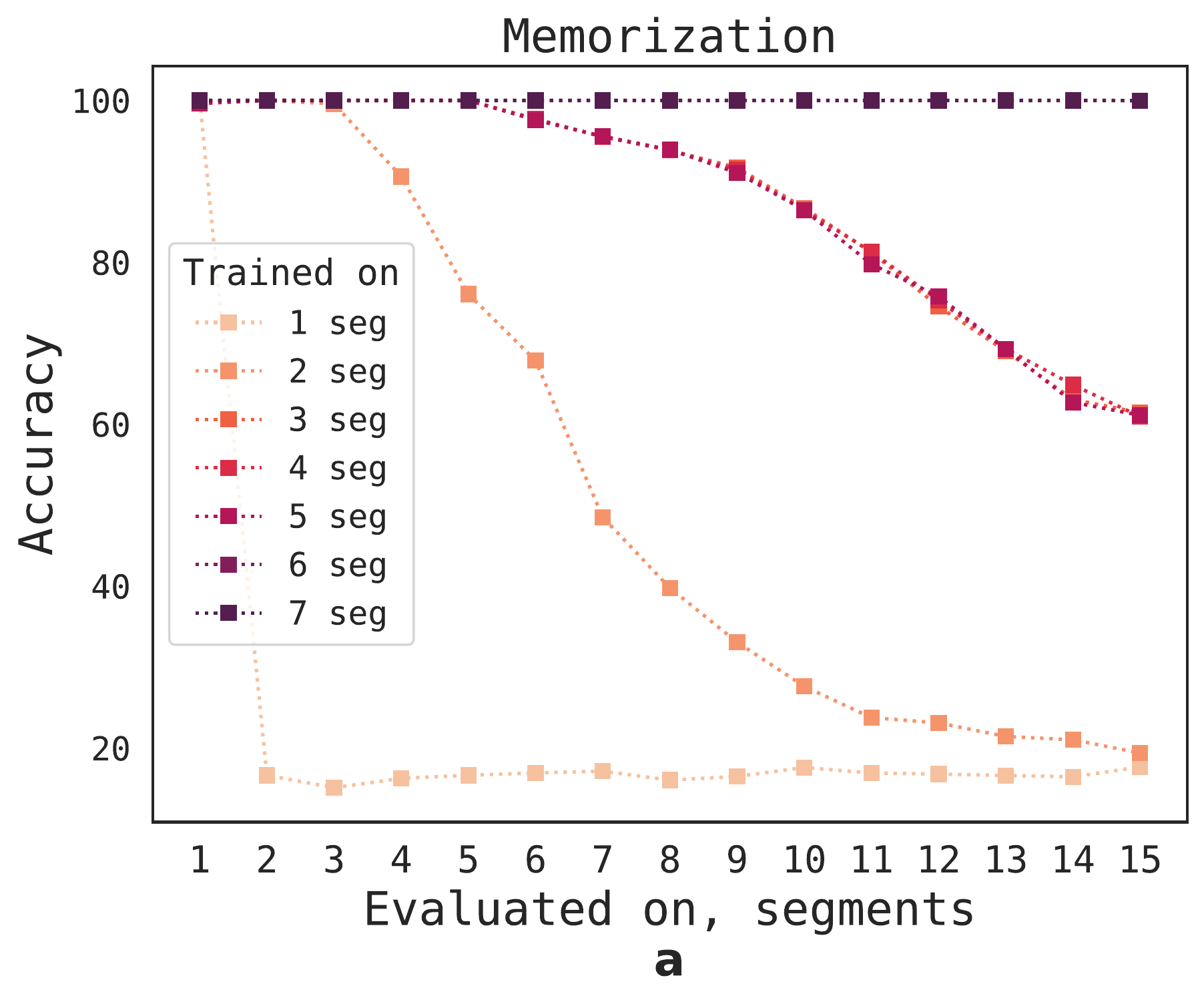}
\includegraphics[width=0.8\linewidth]{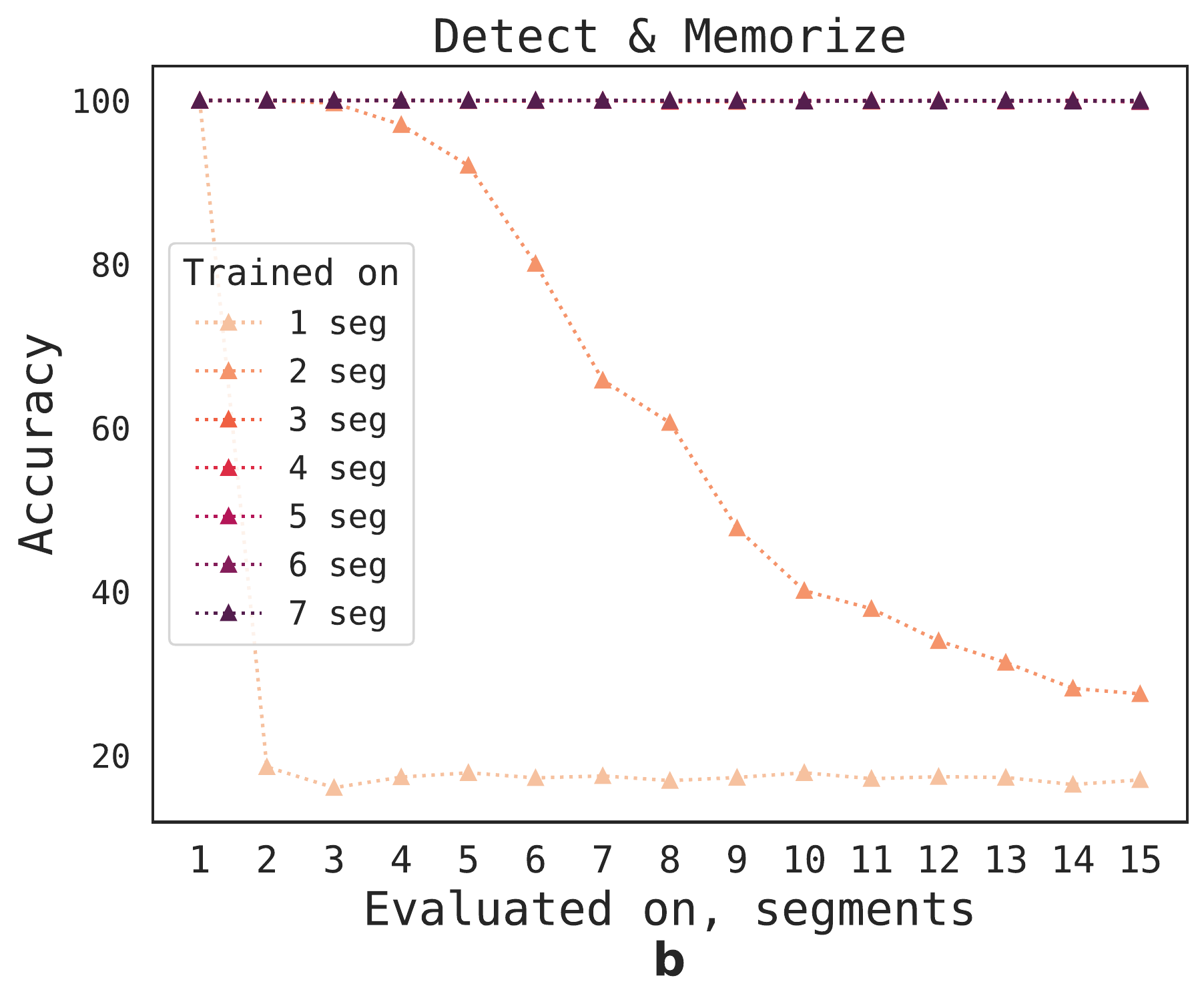} \
\includegraphics[width=0.8\linewidth]{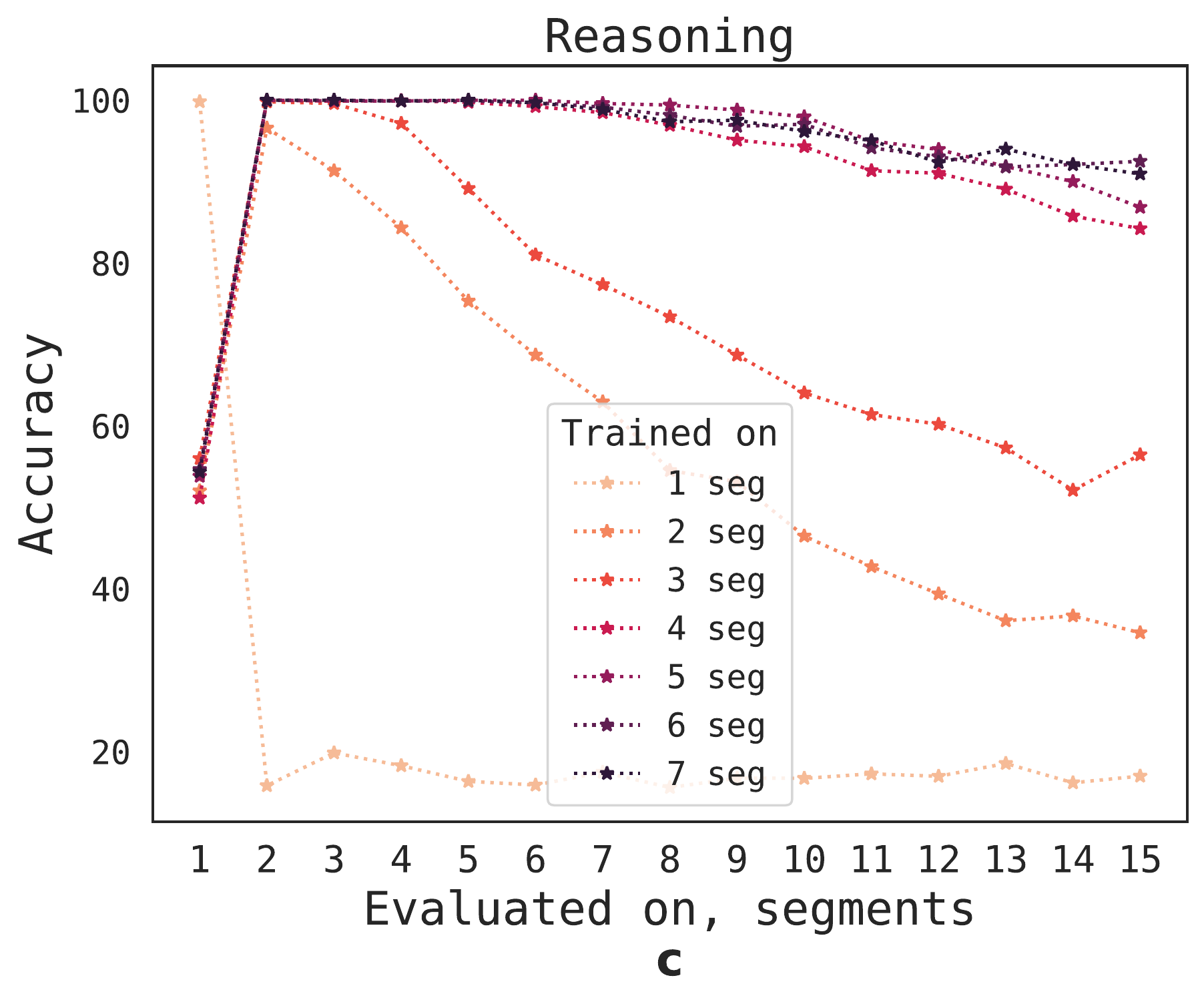}
\caption{\small\textbf{Generalization of memory retrieval}. Evaluation of checkpoints trained on 1-7 segment tasks with memory size 10 on varying input lengths. \textbf{a}: Memorization task, \textbf{b}: Detection \& memorization, \textbf{c}: Reasoning. Models trained on more than 5 segments generalize well on longer tasks.}
\label{extrapolate}
\end{figure}

To improve training stability of the original RMT we introduce curriculum learning. At the beginning of training, RMT is fine-tuned on shortest one segment version of the task, and upon convergence, the task length is increased by adding one more segment. The curriculum learning continues until the desired input length is reached.

\begin{figure*}[!t]
\centering
\includegraphics[width=\linewidth]{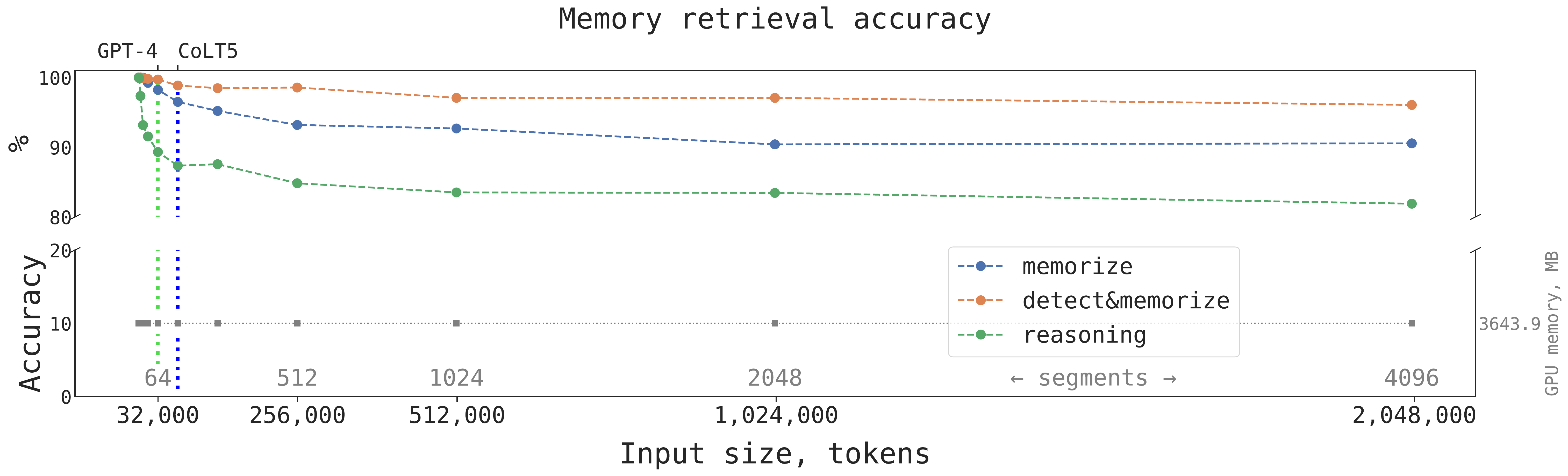}
\caption{\small\textbf{Recurrent Memory Transformer retains information across up to $2\times10^6$ tokens}. By augmenting a pre-trained BERT model with recurrent memory~\citep{rmt_2022}, we enabled it to store task-specific information across 7 segments of 512 tokens each. During inference, the model effectively utilized memory for up to 4,096 segments with a total length of 2,048,000 tokens—significantly exceeding the largest input size reported for transformer models (64K tokens for CoLT5~\citep{ainslie2023colt5}, and 32K tokens for GPT-4~\citep{openai2023gpt4}, and 100K tokens for Claude). This augmentation maintains the base model's memory size at 3.6 GB in our experiments.}
\label{fig:extrapolate-long}
\end{figure*}

In our experiments, we begin with sequences that fit in a single segment. The practical segment size is 499, as 3 special tokens of BERT and 10 placeholders for memory are reserved from the model input, sized 512. We notice that after training on shorter tasks, it is easier for RMT to solve longer versions as it converges faster to the perfect solution.

\begin{figure}
\centering
\includegraphics[width=\linewidth]{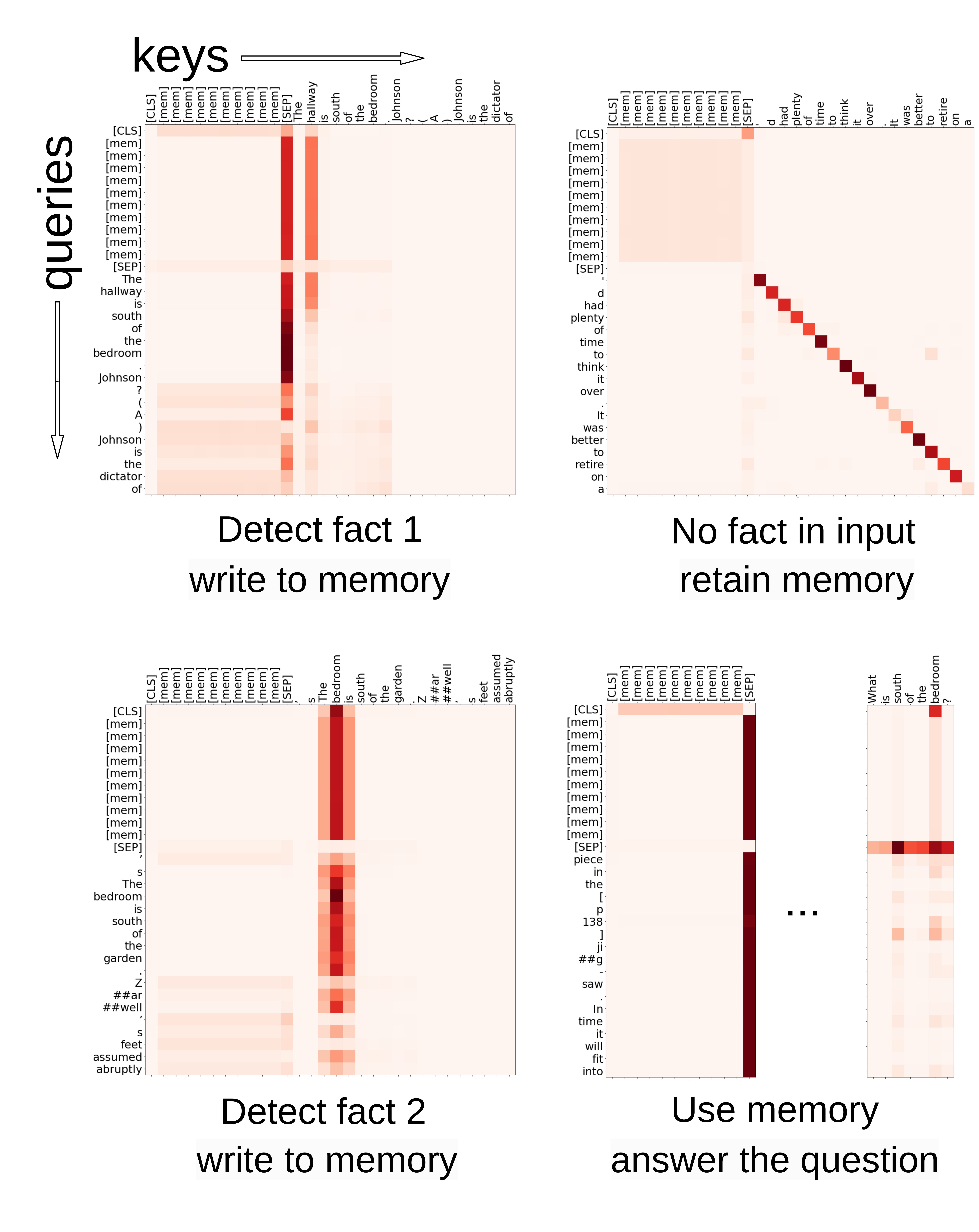}
\caption{\small\textbf{Attention maps for operations with memory}. These heatmaps show operations performed during specific moments of a 4-segment reasoning task. The darkness of each pixel depends on the attention value between the corresponding key and value. From left to right: RMT detects the first fact and writes its content to memory ([mem] tokens); the second segment contains no information, so the memory keeps the content unchanged; RMT detects the second fact in reasoning tasks and appends it to memory; CLS reads information from the memory to answer the question.}
\label{mem_operations}
\end{figure}

How well does RMT generalize to different sequence lengths? To answer this question, we evaluate models trained on a varying number of segments to solve tasks of larger lengths (Figure~\ref{extrapolate}). We observe that most models tend to perform well on shorter tasks. The only exception is the single-segment reasoning task, which becomes hard to solve once the model is trained on longer sequences. One possible explanation is that since the task size exceeds one segment, the model stops expecting the question in the first segment, leading to quality degradation.

Interestingly, the ability of RMT to generalize to longer sequences also emerges with a growing number of training segments. After being trained on 5 or more segments, RMT can generalize nearly perfectly for tasks twice as long. To test the limits of generalization, we increase the validation task size up to 4096 segments or 2,043,904 tokens (Figure~\ref{fig:extrapolate-long}). RMT holds up surprisingly well on such long sequences, with Detect \& Memorize being the easiest and Reasoning task the most complex.

By examining the RMT attention on specific segments, as shown in Figure~\ref{mem_operations}, we observe that memory operations correspond to particular patterns in attention. Furthermore, the high extrapolation performance on extremely long sequences, as presented on the Fig.~\ref{fig:extrapolate-long}, demonstrates the effectiveness of learned memory operations, even when used thousands of times. The RMT does not have any specific memory read/write modules and Transformer learns how to operate with memory recurrently.
This is particularly impressive, considering that these operations were not explicitly motivated by the task loss.

\section{Natural and Formal Language Modeling}

To study the contribution of recurrent memory for long text understanding, we focus on the long range language modeling task. To capture long-term dependencies in text, memory is required to find and store various type of information between segments. We train the GPT-2 Hugging Face checkpoint with 2 memory tokens using the recurrent memory approach on the ArXiv documents from The Pile~\citep{pile}. The dataset is preprocessed by splitting each document into non-overlapping segments of fixed length, which are prepended with their respective histories that consist of several segments. During both training and evaluation we process history and target segments one by one and calculate loss and perplexity only for the last target segment. Similarly to memorization tasks, we employ curriculum learning for training, starting without history and then gradually increasing context size. We also find that mixing the number of segments on each curriculum step leads to much better generalization on other sequence lengths. 
We discuss curriculum procedures and usage of parameter-efficient methods in the Appendix.

\begin{figure}[t]
\centering

\includegraphics[width=0.75\linewidth]{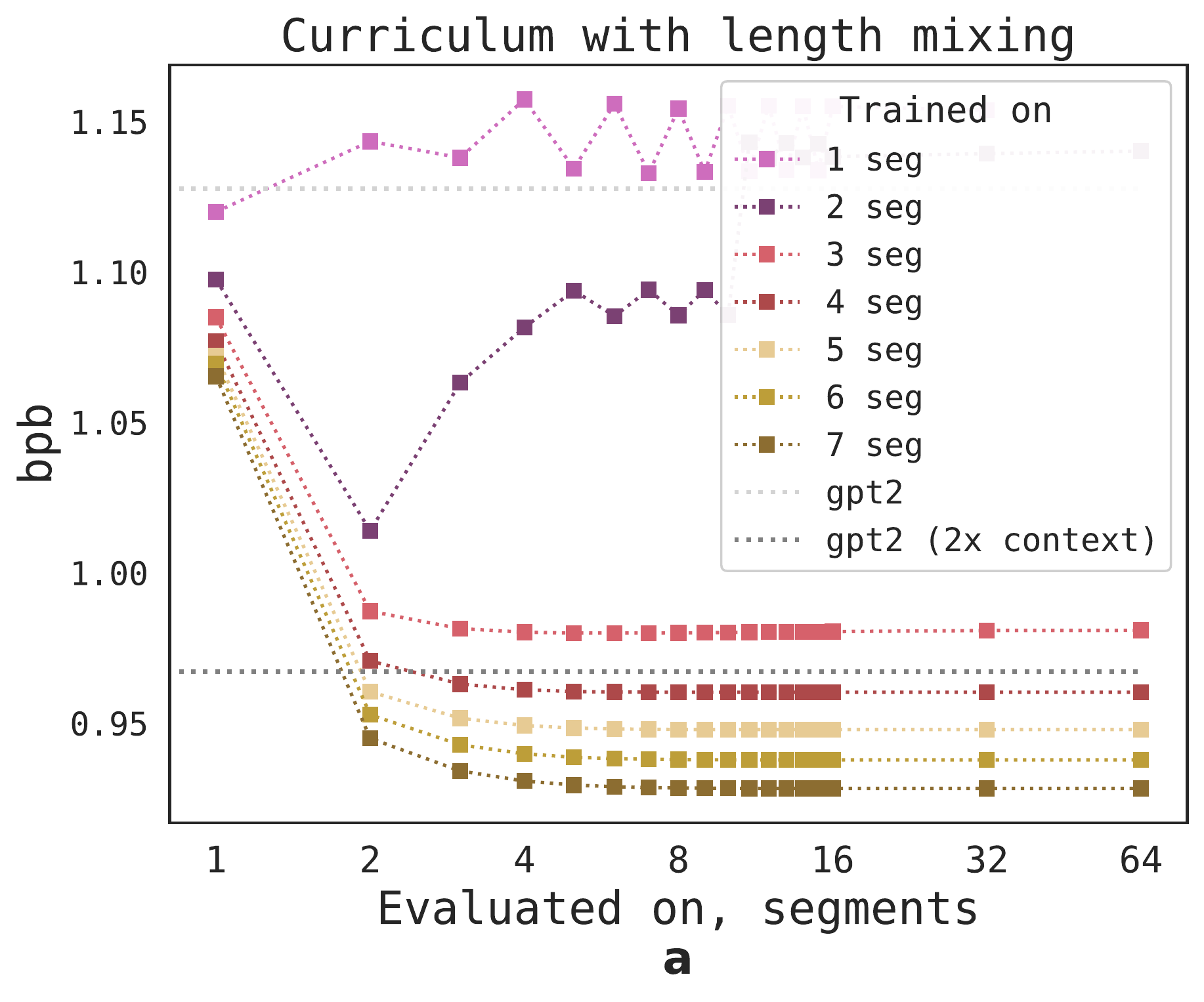}
\includegraphics[width=0.75\linewidth]{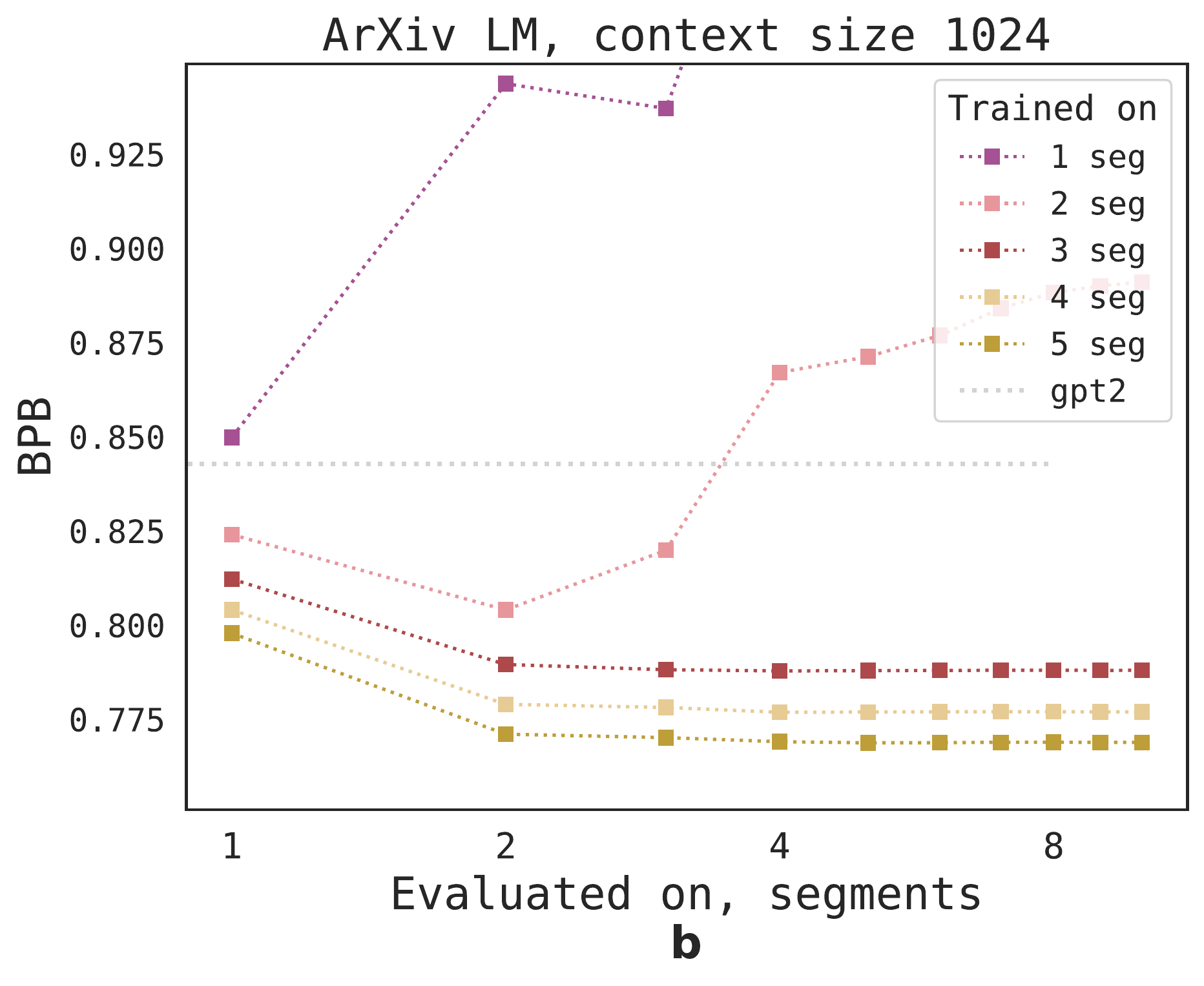}
\caption{\small\textbf{Generalization of memory on language modeling task}.  Models with input sizes \textbf{a:} 128 and \textbf{b:} 1024  trained with RMT show better performance and generalization across longer sizes of context. Perplexity improvement from training RMT with memory size 2 compared to training the baseline GPT-2 for the same number of steps.}
\label{fig:extrapolate_lm}
\end{figure}

As expected, increasing the effective context size leads to an improvement in perplexity (Figure~\ref{fig:extrapolate_lm}). RMT trained for an equal number of steps as the baseline GPT-2 displays substantially lower perplexity values. With increasing number of segments in training RMT starts exhibiting better tolerance to longer history sizes. Performance of memory models trained without history suffers when applied to long contexts, but improves after multi-segment training. 

\begin{figure}[t]
\includegraphics[width=0.9\linewidth]{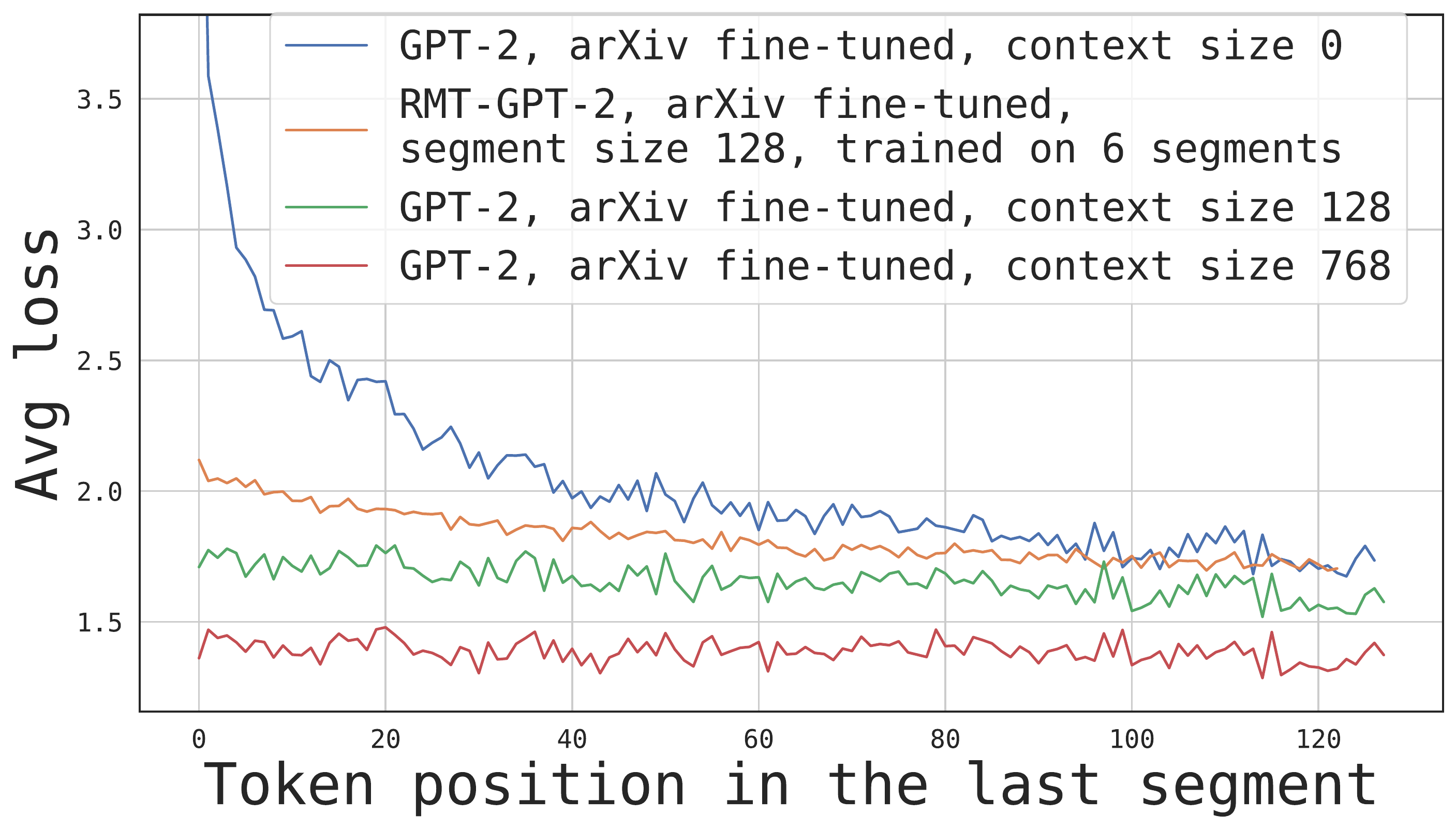}
\caption{\small\textbf{Memory improves prediction at a beginning of a segment.} As we can see, there is an increase in the loss for tokens at the beginning for GPT-2 (context size 0), showing that it struggles to predict the first tokens since they have no context. The RMT keeps information about previous segments in memory tokens, which helps it to improve tokens predictions. However, showing the model the exact previous context (context size 128 and 768) allows for larger loss gains, but at a higher inference cost.}
\label{fig:per_token_loss}
\end{figure}

To understand how memory is utilized during generation of the sequence we measured perplexity for every position in it (see Figure~\ref{fig:per_token_loss}). Baseline shows low prediction quality at the beginning of the sequence due to short context available to condition generation. On the other hand, RMT ensures equally good prediction for all tokens due to carryover of information from the previous segment.

\begin{figure*}[!t]
\centering
\includegraphics[width=0.65\linewidth]{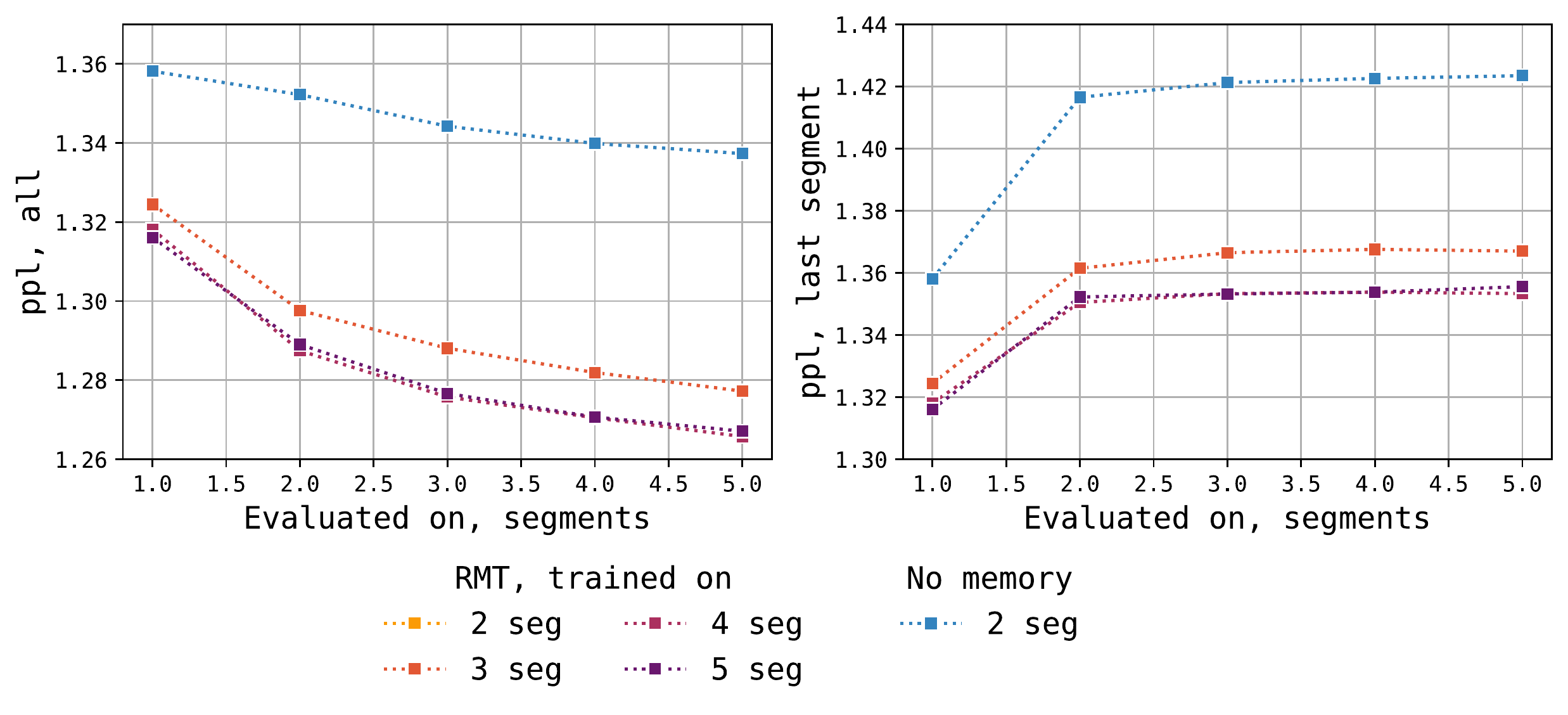}
\caption{\small\textbf{Lemmas memorization for a theorem proving.} Evaluation of the RMT model and backbone model without memory.
Two metrics are calculated: perplexity on all sequence tokens (left) and on the last segment of the sequence (right). RMT model shows better quality.}
\label{fig:proofs}
\end{figure*}

To test our approach in a different domain we fine-tune RMT on a complex mathematical task: generating a proof for a given mathematical theorem in formal language. For our experiments, we utilized Lean 3~\citep{de2015lean} and its library, Mathlib~\citep{The_mathlib_Community_2020}, which contains a range of formalized theories.

Each proof relies on known results, referred to as lemmas. To ensure an effective model, it must accurately assess the relevance of a lemma to the given proof. Subsequently, it should memorize the lemma's name and incorporate it within the proof. To construct our dataset, we organized each sample into a sequence format. The sequence comprises the theorem statement at the beginning, followed by a randomly ordered list of relevant and irrelevant lemmas, and concludes with the human-written proof. By adjusting the presence of irrelevant lemmas, we control the sequence length. We further divide the sequence into non-overlapping segments of fixed size.

For training and evaluation, we calculate the loss and perplexity of the entire sequence. Similar to memorization tasks, we train the RMT model and gradually increase size of the sequences. As our backbone, we employ GPTNeo~\citep{gpt-neo} with 1.3B parameters. We incorporate 10 memory tokens and set the segment size to 2028.

To assess the performance of the RMT model, we compare it with GPTNeo without memory trained on a sequences of 2 segments (first segment always contains the theorem statement and the second contains the proof).
GPTNeo undergoes fine-tuning using the same number of tokens as RMT with 2 segments.
Figure~\ref{fig:proofs} shows the results of the RMT model.
The RMT model improves perplexity compared to the memory-less model.

However, training with 4 or more segments does not enhance predictions for longer sequences.
According to how the sequence is constructed and split into segments,
we hypothesize that the model is more concentrated on learning to remember the beginning of the last lemma in the previous segment to predict its end in the subsequent segment.
The effect of detecting and memorizing relevant lemmas and utilizing them in proof generation is less notable.
We believe that the results can be improved by more careful loss construction and data preparation.

\section{Conclusion}

The problem of long input scaling in Transformers has been extensively studied since the introduction of this architecture. Our research has presented a series of significant advancements in augmenting and training of Transformer language models. The work expands the conventional capabilities of pre-trained encoder-only and decoder-only transformers to an unprecedented level of scalability through the integration of token-based memory storage and segment-level recurrence using recurrent memory (RMT).

We have shown that by employing the RMT combined with curriculum learning, even models pre-trained on shorter sequences can be effectively adapted to manage tasks involving significantly longer sequences. This demonstrates that the input length originally designed for the model does not necessarily restrict its potential capabilities, thus offering a new perspective on the adaptability of Transformers.

Our work further uncovered the remarkable adaptability of the trained RMT models in extrapolating to tasks of varying lengths. The results obtained showcased the RMT's ability to handle sequences exceeding 1 million tokens. Importantly, the computational requirements scaled linearly, thereby maintaining computational efficiency even as task length drastically increased. This is a substantial contribution that could lead to broader applications and improved performance in handling large-scale data. Through an analysis of attention patterns, we provided insight into the operations RMT engages to manipulate memory. 

Overall, our research contributes significantly to the understanding and enhancement of pre-trained Transformer language models. It offers a promising direction for future work, particularly in terms of handling longer sequences and improving the adaptability of these models.

\section*{Limitations and Discussion}

    The curriculum procedure has a substantial impact on the generalization abilities of RMT. Consequently, careful consideration and implementation of curriculum is needed, in contrast to straightforward training of regular Transformers.

    We demonstrate scaling to extremely long sequences such as 2M tokens only on specialized tasks. Unfortunately, there are currently no established benchmarks for NLP tasks with such lengths. However, there are no technical limitations to use the proposed methods on tasks with 2M+ tokens lengths.

    Training with BPTT is less computationally expensive than full attention, but still requires a significant amount of computation. In our experiments, BPTT with a maximum unroll of 7 segments was sufficient to show generalization on much longer sequences. However, larger models would be more expensive to train with BPTT and some tasks may require more segments to generalize. Techniques such as gradient checkpointing, truncated BPTT or parameter efficient training can reduce the amount of required resources.

    Another point is that with unlimited resources and general-purpose information to remember, full attention models might still have an edge in performance.  We can think of full-attention models as an upper bound for RMT, since RMT has to operate only on memory states that represent compressed information, not on actual exact hidden states of the past. Recurrent-based approaches, on the other, hand may be useful in complex step-by-step reasoning tasks, with specialized memory-intensive tasks or in cases where current models are limited~\citep{liu2023lost}.
    
\section*{Acknowledgements}

We are thankful to SberDevices for granting us access to additional computational resources.
A.B. and Y.K.'s work was supported by a grant for research centers in the field of artificial intelligence, provided by the Analytical Center for the Government of the Russian Federation in accordance with the subsidy agreement (agreement identifier 000000D730321P5Q0002) and the agreement with the Moscow Institute of Physics and Technology dated November 1, 2021 No. 70-2021-00138.

\bibliography{aaai24}

\clearpage

\appendix
\setcounter{secnumdepth}{1}

\section{Appendix: Training details}

\subsection{Efficiency of recurrence}

\begin{figure}
    \centering
    \vskip -0.1in
    \includegraphics[width=0.99\linewidth]{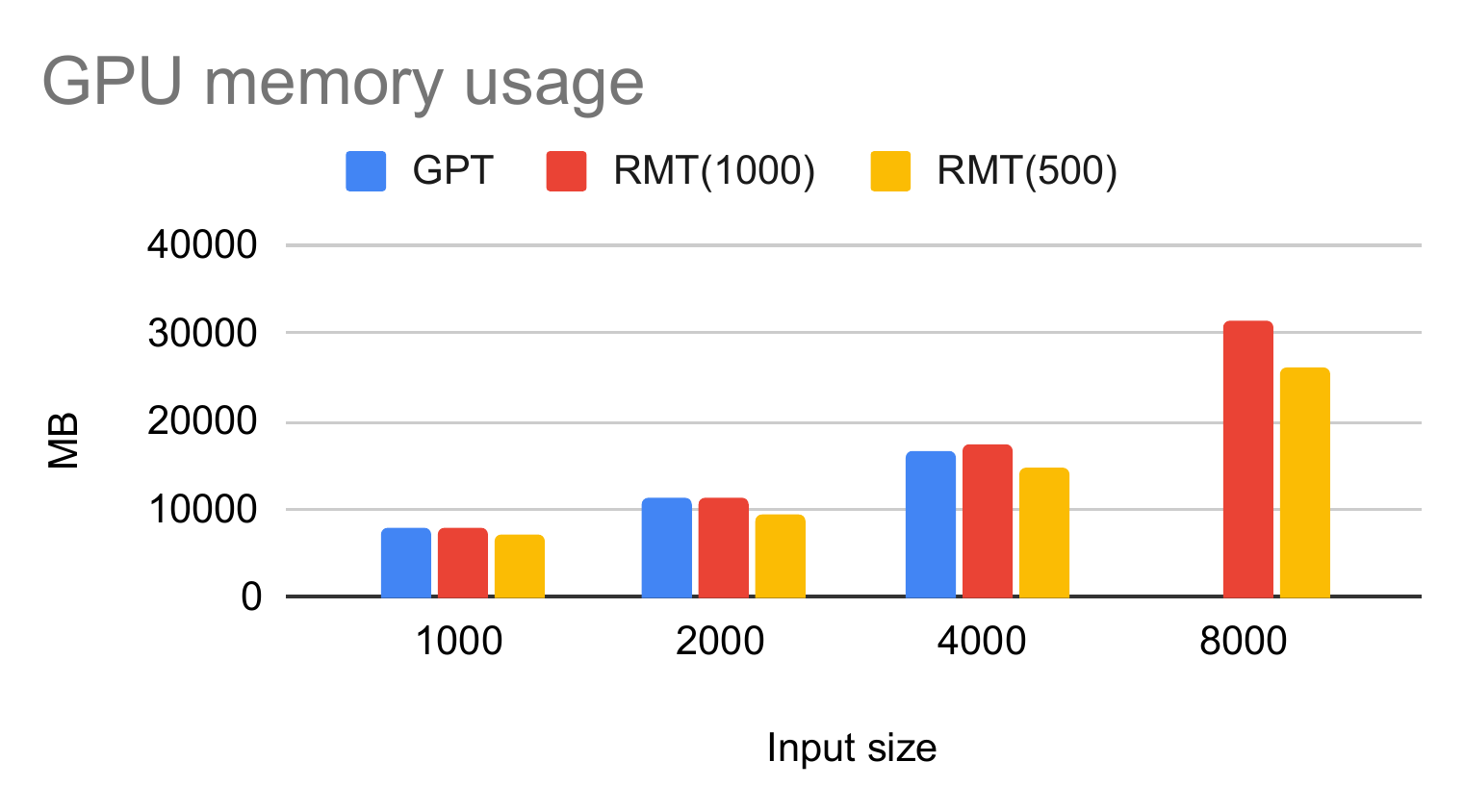}
    \includegraphics[width=0.99\linewidth]{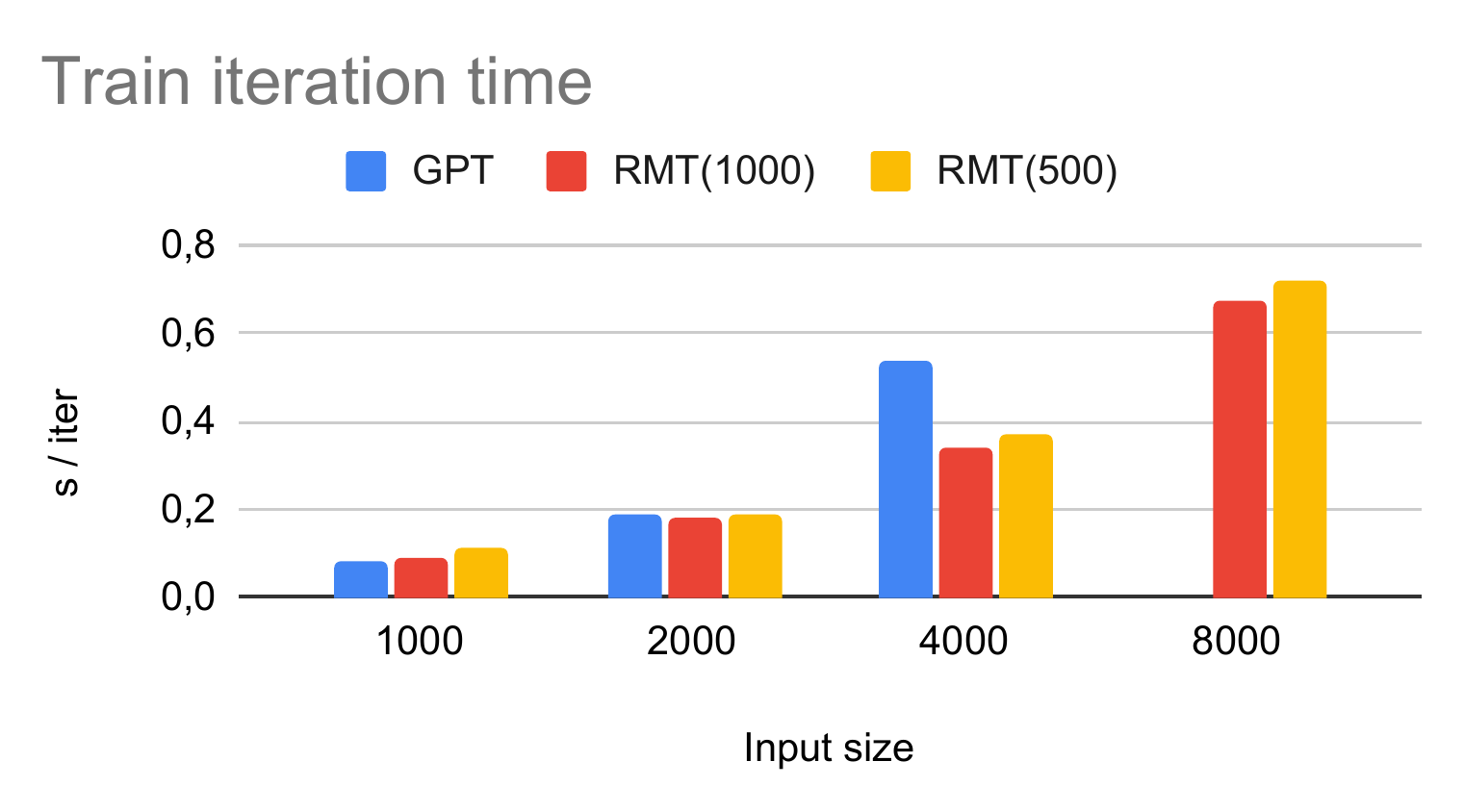}
    \includegraphics[width=0.99\linewidth]{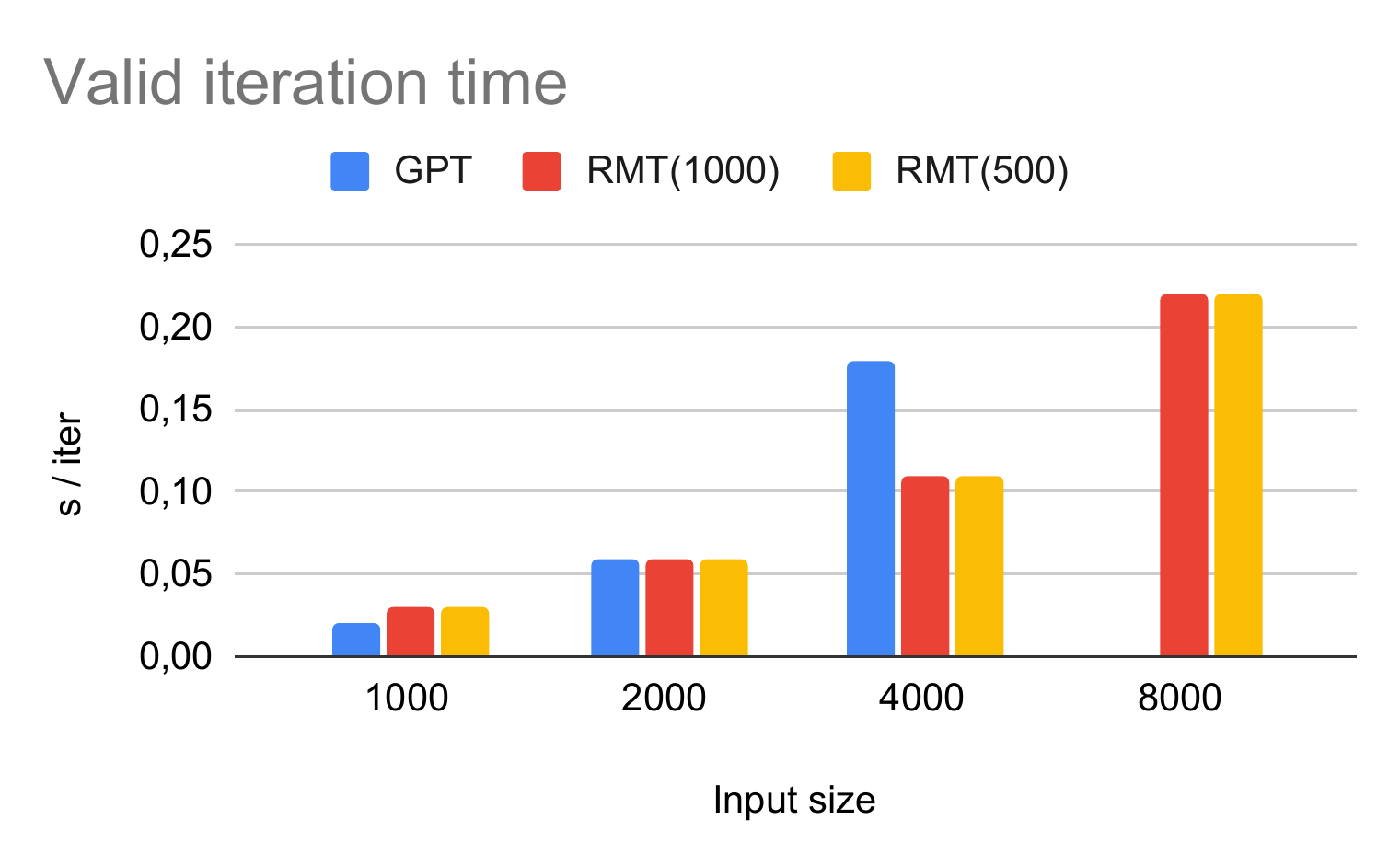}
    \caption{RMT (1000 and 500 segment size) is faster and takes significantly less memory than scaling full attention with input size larger than 4000 tokens. GPT-2 (110M) OOMs on an 80Gb GPU with 8k tokens. In this experiment we do not limit BPTT unroll following the paper pipeline. RMT implementation can be further optimized in terms of speed and memory usage.}
    \label{fig:training_measures}
\end{figure}

We compare the resource efficiency of RMT and full attention model by measuring GPU memory and iteration time on Figure \ref{fig:training_measures}. The tests were done on a single Nvidia A100 80GB GPU. We use reference GPT2 implementation from HuggingFace without FlashAttention and any other optimization techniques, and run models in FP32 mode.

RMT is not only more memory efficient but also faster than the standard transformer, even considering additional overhead from backpropagation through time, that can be turned off when the resources are limited. Baseline transformer fails to process sequences with length 8000 with out-of-memory error. 

It is also worth noting that one of the key advantages of RMT is ability to train on shorter sequences, e.g. up to 8000 tokens and then leverage its generalization capabilities to process much larger sequences. During evaluation, computational requirements for RMT scale linearly, while memory requirements remain constant. This is due to the fact that only one segment is kept in GPU memory at a time during inference.

\subsection{Synthetic task generation}

Here we provide a concise description of memorization tasks. Code for dataset generation and reproducing all experiments can be found in the GitHub repository.

Memorization and Detect\&Memorize datasets use questions and supporting facts from the \textit{qa1\_single-supporting-fact} subset of the bAbI dataset~\cite{WestonBCM15}. 
Facts and questions are constructed using the following pattern: 

\texttt{Fact: [person] [action] [place]}

\texttt{Question: Where is [person]?}

\texttt{Answer: [place] }

The \texttt{[place]} is selected from 6 options: 'bathroom', 'hallway', 'garden', 'office', 'bedroom', and 'kitchen'. For the encoder-only BERT model the resulting task is formulated as a 6-class classification problem, each class being the separate answer option. 

Following the bAbI pipeline, we create a fact and question pair by randomly choosing these options and construct a dataset sample as in Figure \ref{tasks-approaches}. 
In Memorize task, fact is always at the beginning of the input. Detect and Memorize places the fact in a random segment within a sequence. 
Reasoning task is created in a similar way using supporting facts and questions from the \textit{qa4\_two-arg-relations} bAbI subset; an example of such facts is presented at the end of the Memorization Tasks section. 

A noteworthy limitation of the proposed tasks is the source of distractor text, that comes from a different distribution from questions and facts. This makes distinguishing questions trivial for the model. Nonetheless, this could be easily changed to any other texts even from closer distributions. We believe that extending the memorization dataset with complex questions and control over supporting fact position will help improve the way of processing long sequences. Combined with the proposed training schedule, future models can overcome the current limitations of Transformer \citep{liu2023lost}.

\subsection{Impact of curriculum}

In our experiments, we find that the curriculum learning plays an essential role in training RMT. To confirm the importance of the curriculum with a gradually increasing number of segments, we train RMT with and without the curriculum on the memorization task.

\begin{figure}[h]
    \centering
    \includegraphics[width=0.65\linewidth]{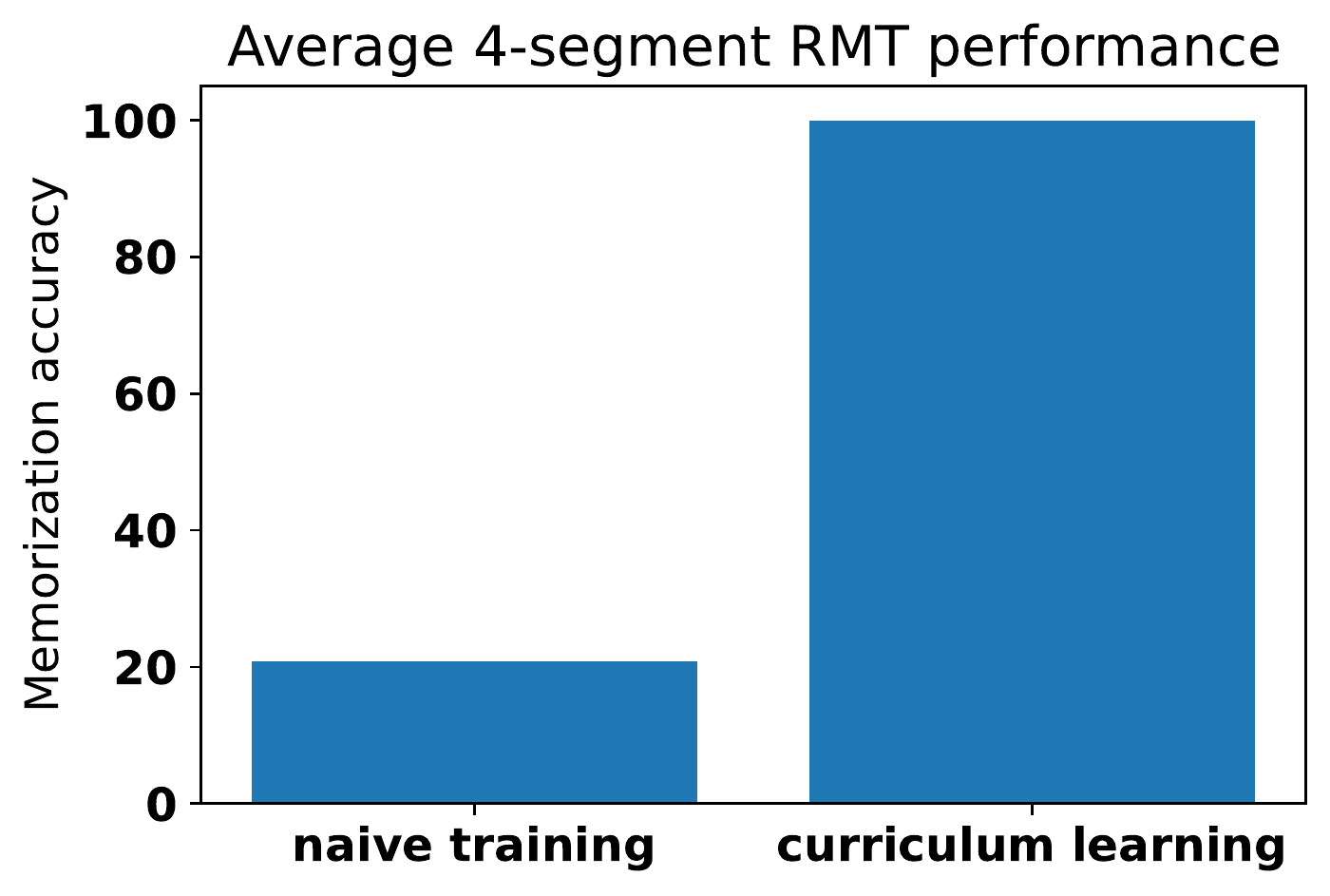}
    \includegraphics[width=0.85\linewidth]{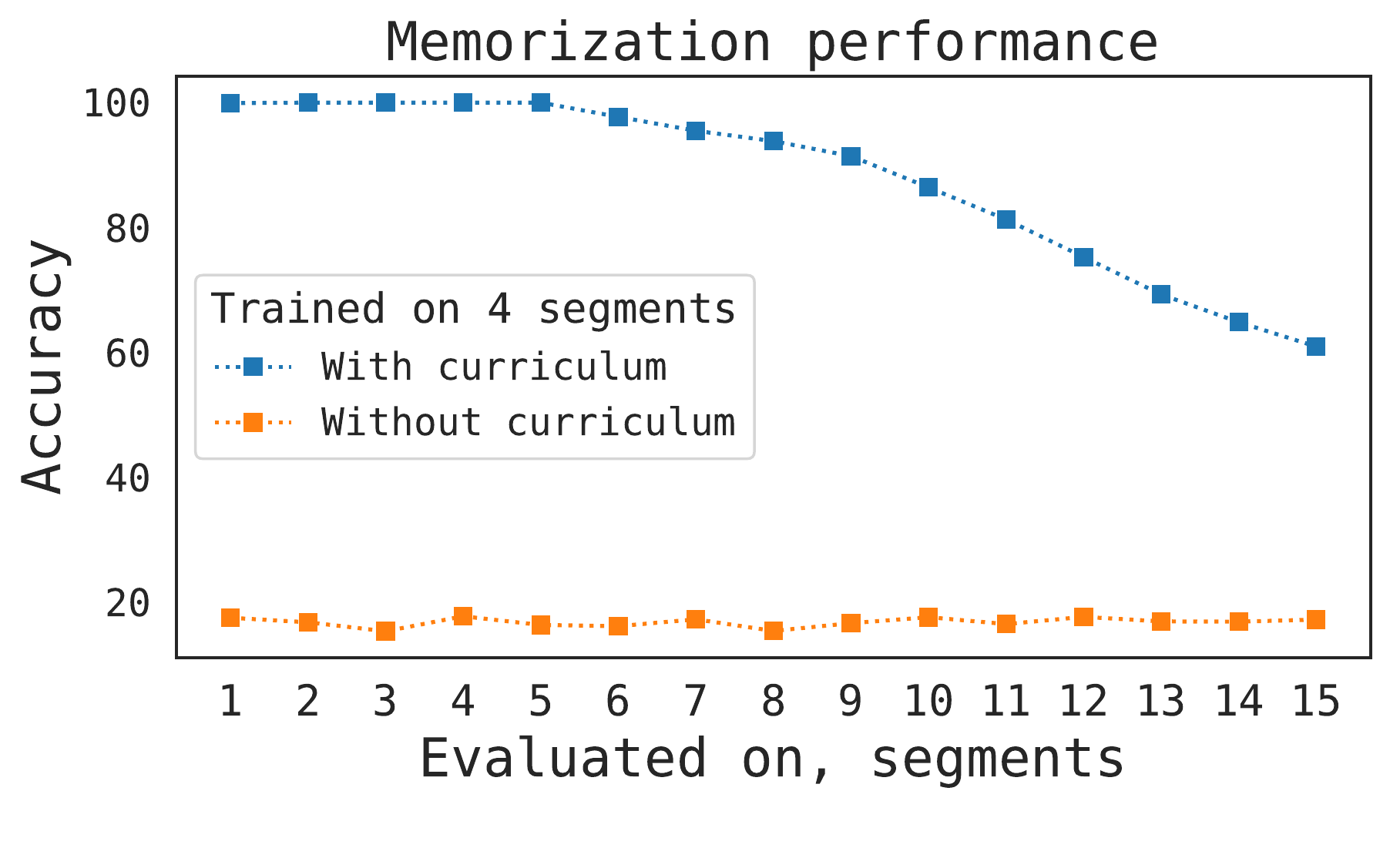}
    \caption{Curriculum boosts RMT abilities to memorize facts and generalize to other sequence lengths. Naive training does not use a curriculum and trains directly on the final maximum number of segments (four in this case).}
    \label{fig:curriculum_impact}
\end{figure}

Figure~\ref{fig:curriculum_impact} shows that in the absence of the curriculum, if the model is trained directly on the maximum number of segments, RMT does not learn neither to solve the task, nor to extrapolate on other sequence length. However, by using the curriculum, a much more capable model with strong generalization capabilities can be obtained.

\begin{figure}
    \centering
    \includegraphics[width=0.9\linewidth]{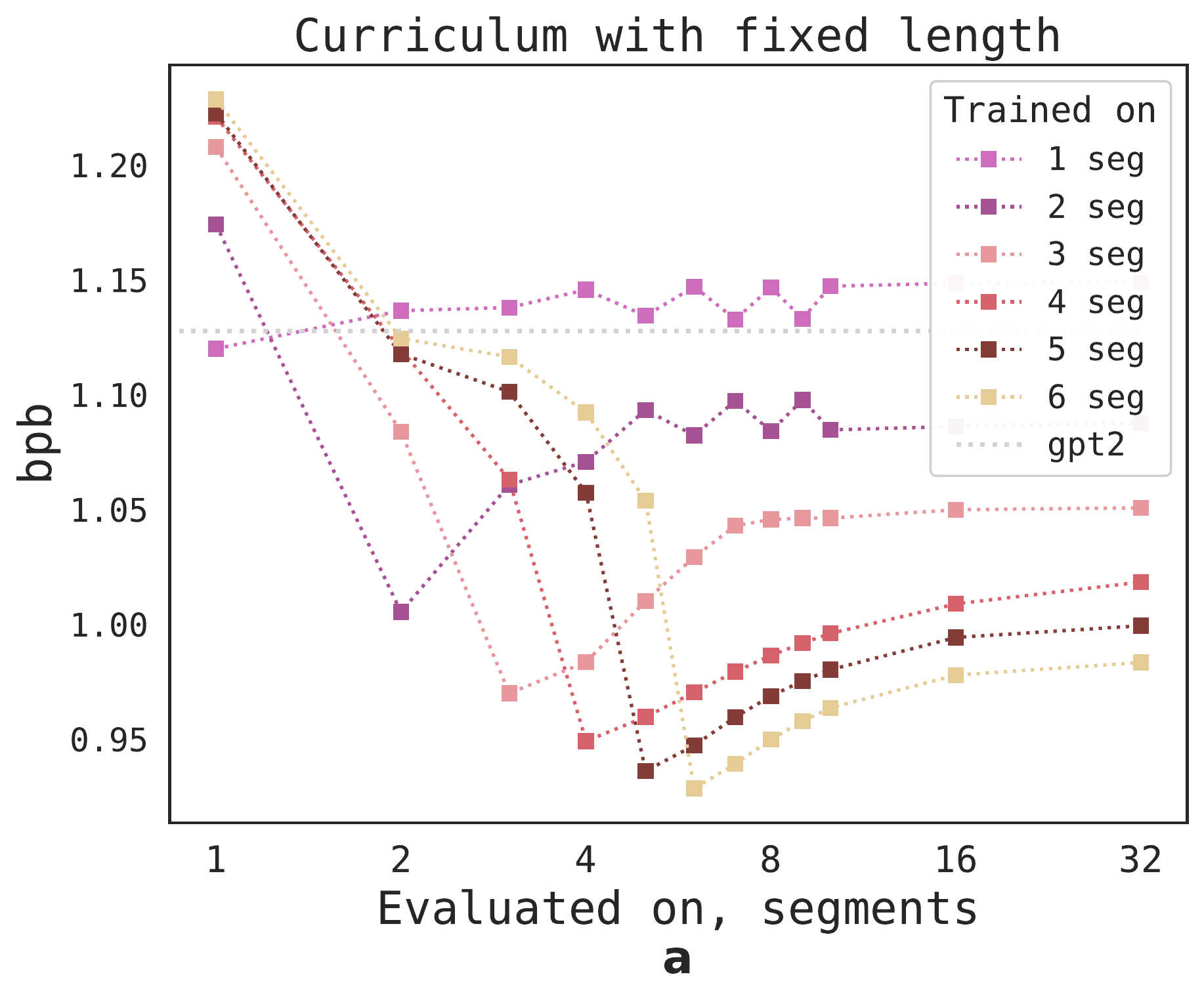}
    \includegraphics[width=0.9\linewidth]{imgs/extrapolate-v2_lm_128_abs.pdf}
    \caption{For Arxiv language modeling task (b) mixing in all previous number of segments during curriculum learning improves generalization compared to (a) using fixed number of segments at each curriculum stage. Curriculum with length mixing improves performance on smaller numbers of segments and shows extrapolation to lengths that were not seen during training. Also, we highlight that RMT with segment size 128 starts to outperform GPT-2 with double size input (256).}
    \label{fig:curriculum_sampling}
\end{figure}

On Figure~\ref{fig:curriculum_sampling} we also show that the curriculum learning procedure can be improved by adding samples from previous curriculum steps, i.e. at the step when we train on $N$ segments, we also add samples with $\leq N$ segments.

\subsection{Usage of Parameter-Efficient Methods}

\begin{table}[h]
\caption{\small RMT can be successfully combined with parameter-efficient methods (parallel adapter, LoRA). Results for language modeling on the Arxiv dataset for Pythia-70m model.}
\label{tab:peft}
\begin{center}
\begin{small}
\begin{sc}
\fontsize{7}{8}
\selectfont 
\begin{tabular}{lc}
\toprule
Model & loss \\
\midrule
Adapter only              &  41.43 \\
Adapter + RMT-1seg        &  10.31 \\
Adapter + LoRA + RMT-1seg &  7.30  \\
Adapter + LoRA + RMT-2seg &  6.97  \\
\bottomrule
\end{tabular}
\end{sc}
\end{small}
\end{center}
\end{table}

One notable advantage of RMT is that the backbone architecture remains unchanged. This makes it possible to utilize existing parameter-efficient methods to modify only a small number of parameters to incorporate memory. The performance of RMT in combination with LoRA~\citep{hu2022lora} and Parallel Adapter~\citep{he2022parallel_adapter} is depicted on Table~\ref{tab:peft}. Adding recurrence results in significant improvement in perplexity for the Pythia model~\citep{biderman2023pythia} compared to using parameter-efficient methods alone.

RMT offers the flexibility of incorporation various cost-efficient training methods, which greatly enhances its practical applicability, especially when computational resources are limited.

\end{document}